# Étiqueter un corpus oral par apprentissage automatique à l'aide de connaissances linguistiques

Iris Eshkol[1], Isabelle Tellier[2], Samer Taalab[2], Sylvie Billot[2]

1 LLL-Université d'Orléans – France
2 LIFO-Université d'Orléans – France
avec l'aide du projet ANR-07-MDCO-03 « CRoTAL »

## Abstract

Thanks to the Eslo1 (« Enquête sociolinguistique d'Orléans », i.e. « Sociolinguistic Inquiery of Orléans) campain, a large oral corpus has been gathered and transcribed in a textual format. The purpose of the work presented here is to associate a morpho-syntactic label to each unit of this corpus. To this aim, we have first studied the specificities of the necessary labels, and their various possible levels of description. This study has led to a new original hierarchical structuration of labels. Then, considering that our new set of labels was different from the one used in every available software, and that these softwares usually do not fit for oral data, we have built a new labeling tool by a Machine Learning approach, from data labeled by Cordial and corrected by hand. We have applied linear CRF (Conditional Random Fields) trying to take the best possible advantage of the linguistic knowledge that was used to define the set of labels. We obtain an accuracy between 85 and 90%, depending of the parameters used.

## Résumé

Grâce à l'"Enquête sociolinguistique d'Orléans" (Eslo1) conduite en 1968, un corpus oral conséquent a été recueilli, puis retranscrit sous une forme textuelle. L'objectif du travail présenté ici est d'associer une étiquette morpho-syntaxique à chaque unité de ce corpus. Pour cela, nous avons tout d'abord mené une réflexion sur la spécificité des étiquettes de l'oral et sur leurs différents niveaux de description possibles. Cette réflexion a abouti à une structuration hiérarchique originale. Ensuite, étant donné que notre jeu d'étiquettes ne coïncidait avec celui d'aucun outil existant, et que ces outils ne sont en général pas adaptés aux données orales, nous avons construit un étiqueteur par apprentissage automatique, à partir de données étiquetées par Cordial et corrigées à la main. Nous avons utilisé des CRF (Conditional Random Fields) linéaires, en essayant d'exploiter au mieux les connaissances linguistiques qui ont présidé à la définition des étiquettes. Nous aboutissons à une correction de 85 à 90% suivant les paramétrages.

**Mots-clés :** corpus oral, étiquetage morpho-syntaxique, apprentissage automatique, CRF

## 1. Introduction

L'étiquetage morpho-syntaxique d'un texte est une étape fondamentale de son analyse, et un préliminaire à tout traitement de plus haut niveau. Des étiqueteurs fiables existent pour le français, mais ils sont conçus pour les textes écrits, et sont de ce fait mal adaptés aux spécificités d'une langue moins « normalisée ». Or le corpus ESLO, auquel nous nous intéressons dans cet article, provient de la transcription d'enregistrements oraux, et présente donc des particularités mal prises en compte par les étiqueteurs standards.

Pour étiqueter l'oral, plusieurs possibilités se présentent : on peut adapter un étiqueteur de l'écrit en lui fournissant des règles formelles qui prennent en compte les disfluences (Dister, 2007), ou adapter la transcription aux exigences de l'écrit (Valli et Véronis, 1999) ou encore développer un étiqueteur spécifique à un corpus donné (Mertens, 2003). Nous avons opté pour une méthodologie différente. Sur la base d'un étiqueteur de l'écrit, nous définissons tout d'abord un jeu d'étiquettes répondant à nos besoins. Nous constituons ensuite un corpus de référence pour ce nouvel étiquetage et nous entraînons avec lui un système d'apprentissage automatique.

Les meilleurs outils actuels capables d'apprendre automatiquement à étiqueter à partir d'exemples sont les CRF (« Conditional Random Fields » ou « Champs Markoviens Conditionnels »). Les CRF sont une famille de modèles statistiques introduits récemment (Lafferty *et al.*, 2001; Sutton et McCallum, 2006), qui ont fait la preuve de leur efficacité dans de nombreuses tâches d'ingénierie linguistique (McCallum et Li, 2003 ; Pinto et *al.*, 2003 : Altun et *al.*, 2003 ; Sha et Pereira, 2003). Pour nos expériences, nous utilisons la bibliothèque libre et gratuite CRF++[1], due à Taku Kado. L'originalité de notre approche est que nous testons diverses stratégies de décomposition des étiquettes en sous-étiquettes plus simples, afin de faciliter l'apprentissage tout en exploitant au mieux les connaissances linguistiques qui ont présidé au choix des étiquettes initiales. Nous suivons en cela la méthodologie de (Jousse 2007; Zidouni et *al* 2009).

Dans la première partie de cet article, nous présentons notre corpus et le processus d'étiquetage, en nous focalisant sur les problèmes qu'il pose pour les corpus oraux. Nous détaillons le choix de notre nouveau jeu d'étiquettes, et la méthode adoptée pour disposer d'un corpus de référence correctement étiqueté. Nous développons ensuite les expériences réalisées avec CRF++ pour apprendre automatiquement un étiqueteur morpho-syntaxique adapté aux spécificités de notre corpus. Nous montrons qu'en jouant sur la décomposition des étiquettes, il est possible d'améliorer l'efficacité de l'apprentissage.

## 2. Un corpus oral et son étiquetage

Cette section est consacrée à l'étiquetage morpho-syntaxique d'un corpus oral, et aux difficultés qu'il pose à un étiqueteur comme Cordial. Les spécificités de l'oral nous amènent à proposer un nouveau jeu d'étiquettes plus adapté que celui de Cordial.

### 2.1. L'étiquetage morpho-syntaxique de l'oral

L'objectif de l'étiquetage que nous cherchons à réaliser est d'attribuer à chacun des mots d'un corpus une étiquette qui récapitule ses informations morpho-syntaxiques. Ce processus d'étiquetage peut s'accompagner de celui de lemmatisation, dont l'objectif est de ramener l'occurrence d'un mot donné à sa forme de base ou « lemme ». La principale difficulté de cet étiquetage est due à l'ambiguïté des mots polycatégoriels (e.g. « portes » est soit le pluriel du nom commun « porte », soit la deuxième personne du singulier du présent de l'indicatif ou du subjonctif du verbe « porter », soit un adjectif comme dans « veine porte ») : un étiqueteur se doit d'attribuer la bonne étiquette dans un contexte donné. Les étiqueteurs doivent aussi faire face à des mots absents des dictionnaires : mots mal orthographiés, noms propres, néologismes, etc.

L'étiquetage d'un corpus oral ajoute des problèmes supplémentaires. Tout d'abord, les transcriptions ne sont en général pas ponctuées pour éviter l'anticipation de l'interprétation (Blanche-Benveniste et Jeanjean, 1987). Les signes de ponctuation comme le point ou la virgule, ainsi que la majuscule au début de l'énoncé, sont des marques typographiques. De même la notion de phrase, essentiellement graphique, a rapidement été abandonnée par les linguistes qui s'intéressent à l'oral. Les études sur la langue parlée ont permis ensuite de dégager des phénomènes propres à l'oral, qu'on regroupe souvent sous l'appellation générale de *disfluences* : répétitions, autocorrections, amorces de mots, etc. En accord avec (Blanche-Benveniste 2005), nous considérons que l'ensemble de ces phénomènes doit être intégré par l'analyse linguistique même s'ils créent des difficultés pour le traitement. Il en va de même d'autres éléments, comme *hein, bon, bien, quoi, voilà, comment dire, etc.* qui apparaissent avec une fréquence élevée dans les corpus oraux et qui, sans ponctuation[2], peuvent être ambigus :

> *il est gentil bien mais / il est bien gentil*

---

[1]  http://crfpp.sourceforge.net/

Les outils actuels d'étiquetage ne sont pas adaptés à l'oral, d'où la difficulté de la tâche.

### *2.2. Présentation du corpus*

L'Enquête SocioLinguistique d'Orléans (ESLO) représente un corpus oral de grande taille : il contient 317 heures de paroles spontanées (4 500 000 mots) et comporte des fiches sur plus de 200 locuteurs. Les situations d'enregistrements sont diverses : des entretiens en face à face, des reprises de contacts informelles comme des discussions entre amis, des enregistrements en micro caché, des interviews de personnalités de la ville (monde politique, syndical, universitaire ou religieux), des conférences ou débats ainsi que des entretiens au Centre Médico Psychopédagogique d'Orléans (entretiens entre une assistante sociale et des parents). Cette enquête, menée entre 1968 et 1971 par des professeurs de français de l'University of Essex (Royaume-Uni) avait pour but de récolter des documents sonores dans une visée didactique[3].

### *2.3. L'étiquetage par Cordial et ses limites*

Le corpus dont nous disposons correspond à 105 fichiers de transcription XML *Transcriber* convertis en fichiers texte, chacun correspondant à une situation d'enregistrement. Les principales conventions de transcription sont l'absence de ponctuation et de majuscule en début d'énoncé ainsi qu'une transcription orthographique normée. La segmentation en « phrases »[4] a été faite soit sur une unité intuitive de type « groupe de souffle » posée par le transcripteur humain, soit sur le tour de parole, défini uniquement par les changements de locuteurs. Afin de disposer d'un corpus étiqueté de référence, les données transcrites ont été soumises à Cordial. Ce logiciel a été choisi pour sa fiabilité. En effet, c'est aujourd'hui un des meilleurs étiqueteurs du français écrit avec une large palette d'étiquettes, riches d'informations linguistiques. L'étiquetage se présente sous la forme de 3 colonnes : mot, lemme et catégorie grammaticale (POS) :

```
comment   comment   ADV
vous      vous      PPER2P
faites    faire     VINDP2P
vous      vous      PPER2P
une       un        DETIFS
omelette  omelette  NCFS
```

Cordial utilise environ 200 étiquettes indiquant les différentes informations morphologiques comme le genre, le nombre ou l'invariabilité pour les noms et les adjectifs ; la distinction en mode, en temps et en personne pour les verbes ; et même la présence du h aspiré au début du mot. Mais, après avoir analysé les résultats de l'étiquetage, un certain nombre d'erreurs ont été perçues. Il s'agit, en premier lieu, des erreurs « classiques » de l'étiquetage comme :

---

[2] Ces mots constituent des énoncés à eux seuls ou se manifestent à différentes places d'un énoncé sans intégrer sa structure (c'est-à-dire sans entrer en relation syntaxique avec un autre élément), ils sont remplacés à l'écrit par des signes de ponctuation.

[3] En 2005, le laboratoire CORAL devenu ensuite LLL (Laboratoire Ligérien de Linguistique) a entrepris de mettre à disposition ce corpus dans le respect des méthodes et des techniques actuelles. Réunis, *ESLO 1* et *ESLO 2* formeront une collection de 700 heures d'enregistrement.

[4] On emploie ce terme pour désigner la présence d'une segmentation, mais elle ne se traduit pas par une ponctuation.

- L'ambiguïté :

    *et vous êtes pour ou contre* (*contre contrer VINDP3S* à la place de *contre contre PREP*[5])

- Les noms propres :

    *les différences qu'il y a entre les lycées les CEG* (*CEG Ceg NPMS* à la place de *CEG CEG NPPIG*[6]) *et les CES* (*CES ce DETDEM* à la place de *CES CES NPPIG*)

- Les locutions :

    *en effet* analysé en deux lignes (*en en PREP* puis *effet effet NCMS*) alors qu'il s'agit d'une locution adverbiale

En second lieu, nous constatons aussi des erreurs propres à la nature orale des données :

- troncation ou amorce : dans les conventions d'ESLO, la séquence amorcée est notée par un tiret, ce qui pose évidemment problème pour l'étiquetage :

    *on fait une ou deux réclam- réclamations* (*réclam- réclamations   réclamréclamations NCMIN*[7])

    au lieu d'analyser cette séquence en deux unités séparées :

    *réclam- reclam- NCI*[8] puis  *réclamation réclamation NCFS*

- interjection : Cordial ne reconnaît pas toutes les interjections présentes dans le corpus oral

    *alors ben* (ben ben NCMIN) *écoutez madame*

    De plus, ce phénomène pose de nouveau le problème de l'ambiguïté car, selon (Dister 2007) « Toute forme peut potentiellement devenir une interjection. On assiste alors à une recatégorisation grammaticale […], le phénomène par lequel un mot ayant une classe grammaticale dans le lexique peut, en discours, changer de classe ». (p. 350).

    *j'ai quand même des attaches euh ben de la campagne qui est proche quoi* (PRI[9])

- répétition et autocorrection :

    *je crois que le* (le le PPER3S au lieu de le le DETDMS) *le* (le le DETDMS) *les saisons*

Il faut noter également un certain nombre d'erreurs provenant de fautes de frappe ou d'orthographe faites par des transcripteurs humains, les transcriptions n'ayant pas été soumises aux correcteurs orthographiques. La correction manuelle d'un fichier étiqueté par Cordial Analyseur a permis d'établir approximativement le taux d'erreur réalisé par le logiciel à 4% .

### 2.4. Nouveau choix d'étiquettes

Afin de mieux adapter l'étiquetage à nos besoins, un certain nombre de modifications ont été apportées au jeu d'étiquettes. D'une part, nous avons essayé, d'« alléger » le nombre d'étiquettes tout en gardant les informations nécessaires, selon nous, à l'analyse linguistique. D'autre part, nous avons été obligés d'adapter les étiquettes à notre corpus et aux conventions de sa transcription. Nous présentons ici une liste (non exhaustive) des modifications :

- De nouvelles étiquettes ont été introduites comme MI (mot inconnu) pour, entre autres, les cas de troncations et PRES (présentateur) pour les tournures comme *il y a, c'est, voilà* très présentes à l'oral ;

---

[5] Les étiquettes de correction proposées ici sont des étiquettes existant dans Cordial.

[6] Nom Propre Pluriel Invariant en Genre

[7] Nom Commun Masculin Invariant en Nombre

[8] Nom Commun Invariable

[9] Pronom Relatif Invariable

- Quelques étiquettes, trop détaillées selon nous dans Cordial, ont été simplifiées. Par exemple, la gamme d'étiquettes concernant les invariances de l'adjectif ou du nom (masculin invariant en nombre, féminin invariant en nombre, singulier invariant en genre, pluriel invariant en genre, invariant en nombre et en genre) a été réduite à une seule étiquette (invariable). Par ailleurs les étiquettes concernant le trait du h aspiré au début du mot ont été supprimées ;
- Afin d'uniformiser le système, certaines étiquettes ont été enrichies : par exemple, les indications sur le genre et le nombre ont été ajoutées aux déterminants démonstratifs et possessifs par souci de cohérence avec d'autres types de déterminants définis ou indéfinis.

Les étiquettes morpho-syntaxiques portent souvent des informations de natures différentes. Elles contiennent toujours l'information sur la partie du discours (POS), encore appelée catégorie grammaticale d'un mot. Mais elles s'enrichissent aussi généralement d'informations :

- morphologiques : concernant la catégorie grammaticale du mot comme son genre, son nombre, l'invariabilité pour les noms, les adjectifs, les déterminants et certains pronoms ;
- syntaxiques : décrivant la fonction du mot dans la phrase et les liens qu'il entretient avec d'autres éléments, comme la mention de coordination et subordination pour les conjonctions ;
- sémantiques : liées à la description du sens des mots comme le caractère possessif, démonstratif, défini, indéfini ou interrogatif pour le déterminant.

Pour rendre compte de ces différentes informations, nous proposons de structurer les étiquettes sur 3 niveaux appelés respectivement L0 (niveau des étiquettes POS), L1 (niveau des variantes morphologiques) et L2 (niveau syntaxico-sémantique), comme dans les exemples ci-dessous :

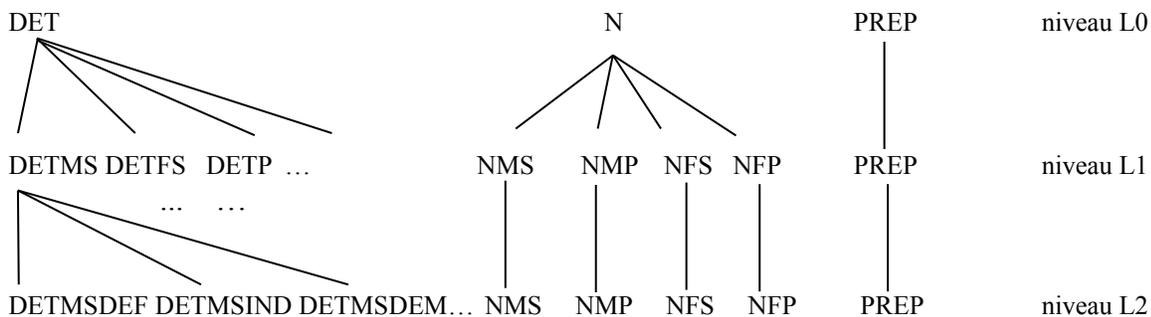

Figure 1: *structuration hiérarchique de quelques étiquettes*

Comme l'illustre la Figure 1, certaines étiquettes :
- restent les mêmes sur les 3 niveaux comme pour les adverbes, les présentateurs, les prépositions, etc. ;
- ne varient qu'au deuxième niveau L1 comme les noms, les adjectifs, les verbes ;
- varient à chaque niveau en intégrant chaque fois de nouvelles informations comme les pronoms et les déterminants.

En plus de cette structuration hiérarchique, d'autres types de connaissances linguistiques peuvent aider à l'étiquetage. Selon la morphologie flexionnelle, qui analyse les mots en constituants, le mot est composé d'une racine et d'une séquence de lettres finales, souvent porteuses de certaines informations morphologiques : des désinences comme -ait, -ais, -is, -é, -s, -s indiquent le temps verbal, le genre et le nombre, etc., c'est ce que la morphologie flexionnelle appelle des morphèmes grammaticaux. En considérant la racine comme la partie commune à toutes les formes d'un mot, il est possible d'extraire ces séquences finales de la forme de surface pour aider à déterminer la partie morphologique de l'étiquette qui doit être associée à ce mot.

Toutes ces connaissances linguistiques peuvent être exploitées pour améliorer les performances d'un système d'apprentissage automatique, comme nous le montrerons dans la section suivante.

Le corpus de référence a été constitué durant le stage de 3 mois d'étudiants linguistes. Il comprend un gros fichier de 18424 mots et 1723 « phrases » (au sens de 2.3). Il a été soumis à Cordial, puis le résultat a été traité à l'aide de scripts et finalement corrigé manuellement afin de le conformer à nos nouvelles conventions d'étiquetage.

## 3. Les expériences

Nous disposons maintenant d'un corpus de référence dont l'étiquetage, validé à la main, est supposé parfait. Il est donc possible de l'utiliser pour entraîner un système d'apprentissage automatique. Le modèle statistique actuellement le plus performant pour apprendre un étiqueteur à partir d'exemples est celui des CRF ou Conditional Random Fields (Lafferty *et al.*, 2001; Sutton et McCallum, 2006). C'est le choix que nous avons fait. Dans cette partie, nous présentons tout d'abord brièvement les propriétés fondamentales des CRF et la façon dont nous avons mené nos expériences, puis nous détaillons leurs résultats. Notre objectif est d'utiliser au maximum les connaissances linguistiques qui ont guidé la définition des étiquettes pour améliorer la qualité de l'étiqueteur appris automatiquement. Nous essayons notamment de voir si l'apprentissage direct des étiquettes ayant tous les niveaux d'information peut être amélioré par une succession d'apprentissages intermédiaires de niveaux d'information moins précis. Nous ne disposons, en revanche, d'aucun dictionnaire énumérant les étiquettes possibles d'une unité textuelle.

### *3.1. CRF et CRF++*

Les CRF sont une famille de modèles statistiques qui permettent d'associer à une observation x une annotation y, en se basant sur un ensemble d'exemples étiquetés, c'est-à-dire un ensemble de couples (x,y). Dans notre cas, chaque x coïncide avec une séquence de mots, éventuellement enrichis d'informations supplémentaires (par exemple si les lemmes correspondant aux mots sont disponibles, x devient une séquence de couples (mot, lemme)) et y est la séquence des étiquettes morpho-syntaxiques associées. Rappelons que pour le corpus oral dont nous disposons, les seuls séparateurs de « phrases », et donc de séquences x, sont dus soit à une pause prolongée notée par le transcripteur manuel, soit à un changement de tour de parole.

Dans un CRF, à la fois x et y sont décomposés en *variables aléatoires* qui ont pour valeurs possibles respectivement les mots (éventuellement enrichis) pour x, et les étiquettes pour y. Il existe autant de variables aléatoires $X_i$ et $Y_i$ qu'il y a de positions possibles i dans une séquence, donc autant que le nombre de mots de la plus longue « phrase » du corpus. Les dépendances entre les variables aléatoires $Y_i$ sont représentées dans un graphe non orienté. L'hypothèse fondamentale sous-jacente est que la valeur d'une étiquette $Y_i$ ne dépend que de la valeur des étiquettes dans la ou les clique(s) (i.e. les sous-graphes complètement connectés) du graphe dont $Y_i$ fait partie, et de la valeur de *n'importe quelle autre* information accessible dans les autres variables $X_j$ présentes dans l'observation. Ce modèle est potentiellement plus riche que celui des HMM (« Hidden Markov Models » ou « Chaînes de Markov Cachées »), et il donne en général de meilleurs résultats.

Dans les CRF linéaires, adaptés à l'annotation de séquences, le graphe relie simplement entre elles les variables d'annotation successives associées aux éléments de la séquence. Les cliques maximales de ce type de graphes sont donc les couples ($Y_i$, $Y_{i+1}$) d'annotations successives. Pour rendre compte de ces dépendances, les CRF font appel à un ensemble de « fonctions features » dont les paramètres sont la clique considérée, les valeurs des variables $Y_i$ à l'intérieur de cette clique, et les variables $X_j$ n'importe où dans la même séquence.

Le logiciel CRF++, que nous utilisons, est fondé sur ce modèle. Les fonctions features y sont définies à l'aide de « templates » ou « patrons » instanciés grâce aux exemples (x,y) fournis au programme. Nous avons conservé les patrons par défaut du logiciel, qui génèrent des fonctions booléennes tenant compte des mots situés dans un voisinage de 2 autour de la position courante.

*Exemple :*   Notre exemple étiqueté (x,y) est celui donné au début de la section 2.3., où la première colonne correspond à l'observation x, la troisième à l'annotation y. Ainsi :

x= c*omment vous faites vous une omelette,*  y= ADV PPER2P VINDP2 PPER2P DETIFS NCFS.

Le patron de génération des features teste, pour une position i donnée identifiant la clique (i, i+1), la valeur des Y dans la clique et la valeur des X en position i, i-2, i-1, i+1 et i+2.  Pour notre exemple et pour la position i=3, on obtient donc la feature suivante :

Si $Y_i$=VINDP2 et $Y_{i+1}$=PPER2P et $X_i$=faites et $X_{i-2}$=comment et $X_{i-1}$=vous et $X_{i+1}$=vous et $X_{i+2}$=une alors la fonction vaut 1 sinon elle vaut 0.

Le patron génère aussi les features plus simples, où seules les positions i, i-1 et i+1 de X sont testées, par exemple. On voit sur cet exemple que les disfluences sont prises en compte dans le modèle directement par le fait qu'elles apparaissent dans les exemples.

Dans la phase d'apprentissage à partir d'exemples (x,y), toutes les fonctions features possibles sont générées et le logiciel associe à chacune d'elle un poids qui optimise la vraisemblance de l'étiquetage proposé. Une fois cette phase d'apprentissage réalisée, quand on fournit au système une nouvelle séquence x non étiquetée, il est capable de proposer l'étiquetage y le plus probable relativement au modèle (et donc aux poids) qu'il a précédemment appris.

*3.2. Cadre des expériences*

Pour valider nos expériences, nous avons systématiquement procédé à une validation croisée : le corpus étiqueté est partagé en dix ensembles, chacun d'entre eux servant successivement d'ensemble de test (dans ce cas, les étiquettes initiales sont bien entendu retirées) après apprentissage à partir des neuf autres ensembles étiquetés. L'étiquetage obtenu sur l'ensemble de test est alors comparé à l'étiquetage exact attendu, ce qui permet de calculer la correction (« accuracy ») de l'étiquetage appris. Les résultats que nous fournissons sont donc toujours la moyenne de 10 apprentissages différents, évalués sur chacun de ces ensembles de tests.

Les fonctions features permettant l'apprentissage sont principalement construites à partir de l'observation des mots. Nous avons aussi réalisé des expériences où le lemme correspondant est supposé connu. Pour rendre plus riches encore les données, nous utilisons aussi les connaissances inspirées de la morphologie flexionnelles, évoquées en section 2.4. :

1. Les notions de « racine » et de « reste » : la racine est la partie de chaîne de caractères commune au mot et au lemme, le reste est ce qui diffère entre eux.  Si mot = lemme, on notera par convention Rmot = Rlemme = 'x ', sinon : mot = Racine + Rmot (où + désigne ici la concaténation de chaînes de caractères) et lemme = Racine + Rlemme.
2. La notion de « dernières lettres d'un mot » : $D_n$(mot) = n dernières lettres de mot.

Par exemple : si mot = 'marchant' et lemme = 'marcher', alors Racine = 'march', Rmot = 'ant', Rlemme = 'er' et $D_2$(mot) = 'nt'.

### 3.3. Expériences de référence

Les expériences de référence sont celles qui consistent à essayer d'apprendre directement le niveau le plus précis (L2), celui où les étiquettes contiennent le plus d'informations possibles. Nous avons d'abord réalisé les tests suivants :

- **Test I** : les fonctions features sont construites à partir de *mot, lemme.* On en produit ainsi 10 000 000 environ, les étiquettes obtenues sont correctes à 86%.
- **Test II :** les features sont construites à partir de *mot, lemme, Rmot, Rlemme.* 11 000 000 sont générées, on atteint 88% de correction.
- **Test III** : Si mot = lemme, on utilise $D_2(mot)$ et $D_3(lemme)$. Les features sont donc construites à partir de *mot, lemme, Rmot|$D_2(mot)$, Rlemme|$D_3(lemme)$*. 20 000 000 features environs sont générées, mais on n'atteint que 82% de correction.
- **Test IV** : comme **III**, mais en utilisant $D_3$ partout. Les features sont donc construites à partir de *mot, lemme, Rmot|$D_3(mot)$, Rlemme|$D_3(lemme)$*. Cette fois, avec le même nombre de features que précédemment, on a 89% de correction, le meilleur taux que nous ayons réussi à obtenir directement.

Comme on pouvait s'y attendre, plus les features sont construites sur des informations riches, meilleurs sont les résultats de l'apprentissage. Si les lemmes ne sont pas supposés connus, nous obtenons les résultats suivants :

- **Test IIIbis** : les features sont construites à partir de *mot, $D_3(mot)$*. Dans ce cas, 8 000 000 sont produites, et le taux de correction est de 87%.
- **Test IVbis** : les features sont construites à partir de *mot, $D_3(mot)$, $D_2(mot)$, $D_1(mot)$*. Cela produit 20 000 000 features environ, la correction atteint 88%.

La connaissance du *lemme* apporte donc en moyenne 2 points de correction, mais il se paie par un apprentissage plus long dû à un plus grand nombre de fonctions features générées.

### 3.4. Apprentissage en cascade

Pour exploiter les connaissances que nous avons sur les étiquettes c'est-à-dire principalement leur structuration en 3 niveaux de hiérarchie, nous avons d'abord essayé d'apprendre chacun des niveaux indépendamment par les mêmes **Tests I**, **II**, **III**, **IV**, **IIIbis** et **IVbis** que précédemment. Nous obtenons alors les résultats du Tableau 1.

| Niveau (nb étiquettes) | Test I | Test II | Test III | Test IV | Test IIIbis | Test IVbis |
|---|---|---|---|---|---|---|
| L0 (16) | 93 | 93 | 94 | 94 | 92 | 93 |
| L1 (72) | 86 | 89 | 90 | 90 | 88 | 89 |
| L2 (107) | 86 | 88 | 82 | 89 | 87 | 88 |

*Tableau 1 : résultats de l'apprentissage de chacun des niveaux d'étiquettes par différents tests*

On peut voir que plus les niveaux sont simples (en termes de richesse d'informations) plus ils sont faciles à apprendre, grâce surtout au nombre plus réduit d'étiquettes. Or, d'après leur organisation hiérarchique (cf. section 2.4.) chaque niveau dépend du précédent : on peut donc espérer améliorer l'apprentissage d'un niveau $L_i$ en utilisant les résultats des niveaux $L_j$, pour j<i, appris précédemment. C'est ce que nous avons fait dans les expériences suivantes, où nous avons appris les niveaux hiérarchiques en *cascade,* à la façon de (Jousse 2007 ; Zidounie et *al* 2009). Nous avons cherché à améliorer les tests précédents qui se comportent le mieux au niveau L0, à savoir les Tests **III** et et **IV.** Les résultats sont présentés en Figure 2.

**Test V :** Ce test est dérivé du **Test III**. Les *mots*, *lemmes* et $D_3(lemme)$ servent à générer les features permettant d'apprendre le niveau L0. Puis le résultat RL0 est utilisé avec les mêmes données pour apprendre le niveau L1, etc. Les apprentissages successifs sont ainsi les suivants :

- CRF (but : L0 | feature (*mot*, *lemme*, $D_3(lemme)$ ) ) → ResL0
- CRF (but : L1 | feature (*mot*, *lemme*, $D_3(lemme)$, *ResL0*)) → ResL01
- CRF (but : L2 | feature (*mot*, *lemme*, $D_3(lemme)$, *ResL0, ResL01*)) → ResL012

**Test VI** : Ce test est dérivé du **Test IV.** Nous générons cette fois les features initiales avec *mot, Rmot, Rlemme, $D_3(mot)$, $D_3(lemme)$*, avec la succession d'apprentissages suivants :

- CRF (but : L0 | feature (*mot, Rmot, Rlemme, $D_3(mot)$, $D_3(lemme)$*)) → ResL0
- CRF (but : L1 | feature (*mot, Rmot, Rlemme, $D_3(mot)$, $D_3(lemme)$*), *ResL0*) → ResL01
- CRF (but : L2 | feature (*mot, Rmot, Rlemme, $D_3(mot)$, $D_3(lemme)$*), *ResL0, ResL01*)) → ResL012

| niveau | IV | VI | III | V |
|---|---|---|---|---|
| **L0** | 94 | 94 | 94 | 94 |
| **L1** | 90 | 90 | 90 | 88 |
| **L2** | 89 | 89 | 82 | 87 |

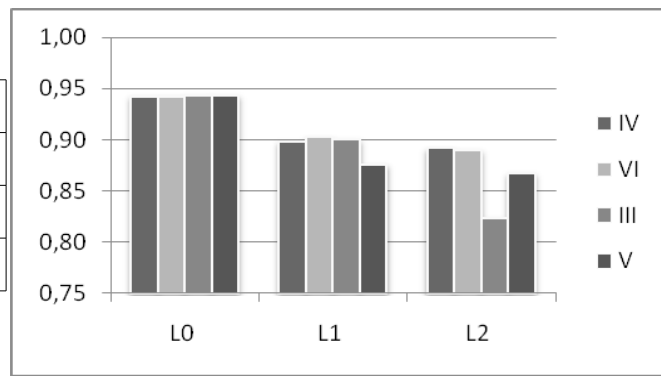

*Figure 2 : Résultats des tests III, IV, V et VI*

Les **Tests V** et **VI** donnent de bons résultats, mais pas vraiment meilleurs que les tests initiaux **III** et **IV**. Malheureusement, l'apprentissage en cascade ne semble donc pas vraiment améliorer les résultats obtenus dans les expériences de référence, où l'apprentissage du niveau L2 se fait directement. Ces conclusions sont confirmées par les expériences sans les lemmes, dont nous ne détaillons pas les résultats. Il faut donc considérer une autre manière de décomposer les informations contenues dans les étiquettes de niveau L2 pour espérer les apprendre mieux.

### *3.5. Apprentissage par décomposition et recomposition d'étiquettes*

A la place des niveaux hiérarchiques L0 L1 et L2, nous avons construit des groupes d'informations à partir desquels il est possible de reconstruire les étiquettes de niveau L2. Notre objectif est d'avoir des groupes *indépendants* (qui concernent un seul type d'information ou des types compatibles), *réduits* (en termes d'effectifs) et tels que la recomposition des vraies étiquettes produise le moins possible d'étiquettes fausses (n'ayant pas de sens linguistique).

**Exemple :** Les étiquettes NFS, NFP, NMS, NMP sont issues de la concaténation des éléments des ensembles {N} (pour la catégorie des noms communs), {M, F} (pour le genre) et {S, P} (pour le nombre). On peut en effet reproduire ces quatre étiquettes par produit cartésien des trois composantes : {N}.{M, F}.{S, P} avec la concaténation « . » comme opérateur entre sous-étiquettes. C'est ce genre de décomposition que l'on souhaite généraliser.

10	Iris Eshkol, Isabelle Tellier, Samer Taalab, Sylvie Billot**Définition :** On appelle « *composante* d'étiquettes » un ensemble de symboles (ou atomes) mutuellement exclusifs à l'intérieur d'une même étiquette. Ces composantes correspondent souvent aux différentes valeurs possibles d'un « trait linguistique » comme le genre ou le nombre. Une solution possible pour construire les composantes de l'ensemble des étiquettes de niveau L2, le plus détaillé de l'arbre hiérarchique, est de prendre les ensembles suivants :

- POS={ADJ,ADV,CH,CONJCOO, CONJSUB, DET, INT, MI, N, PREP, PRES, P, PP, V}
- Genre={M, F} ; Pers={1, 2, 3} ; Nombre={S, P}
- Mode_Temps={CON, IMP, SUB, IND, INDF, INDI, INDP, INF, PARP, PARPRES}
- Dét_Pro={IND, DEM, DEF, POSS, PER, INT}

Il est cependant possible de regrouper encore certaines composantes mutuellement exclusives : par exemple Personne et Genre peuvent être associées parce que '1','2' ou '3' ne figurent jamais avec 'M' et 'F'. Par contre, on ne peut pas regrouper Genre et Nombre parce que 'F' figure avec 'S' ou 'P' dans une même étiquette. Nous proposons finalement les composantes suivantes :

**G0** = POS, **G1** = Genre $\cup$ Pers $\cup \{\varepsilon\}$, **G2** = Nb $\cup \{\varepsilon\}$, **G3** = Mode_Temps $\cup$ Dét_Pro $\cup\{\varepsilon\}$.

$\varepsilon$ étant la chaîne vide, élément neutre pour la concaténation.

Chaque groupe d'étiquettes **Gi** peut être appris indépendamment, par un CRF différent. L'étiquette finale apprise sera alors obtenue par concaténation du résultat de chaque CRF. Le produit cartésien **G0**.**G1**.**G2**.**G3** permet effectivement de générer toutes les étiquettes de niveau L2, et même un peu plus. Par exemple : ADVMP=ADV.M.P. $\varepsilon$ n'a aucun sens car les adverbes sont invariables. Pour résoudre ce problème d'étiquettes produites mais linguistiquement incorrectes, nous avons testé deux méthodes différentes. La première consiste à utiliser un nouveau CRF dont les features sont les composantes apprises indépendamment. La seconde consiste à introduire des règles symboliques explicites au moment de réaliser la concaténation entre sous-étiquettes. Les règles symboliques introduites sont par exemple :

- ADV, CONJCOO, CONJSUB et INT ne peuvent se composer qu'avec $\varepsilon$.
- V ne peut pas se composer avec les symboles appartenant aux composantes Dét_Pro
- DET ne peut pas se composer avec les symboles appartenant à Mode_Temps

Ces règles exploitent le fait que la catégorie POS (composante **G0**, correspondant aussi avec le niveau L0) est apprise avec suffisamment de confiance (94% de correction dans nos expériences) pour contraindre les autres sous-étiquettes avec lesquelles elle est susceptible de se combiner.

Pour apprendre les composantes **G0**, **G1**, **G2** et **G3**, puis la recomposition des étiquettes, nous avons donc réalisé les tests suivants, pour i valant de 0 à 3 :

CRF (but : Gi | feature (*mot, lemme, $D_3(mot)$*)) $\rightarrow$ ResGi

Nous avons aussi testé les variantes où la génération des fonctions features se fait sans lemme, avec à la place les terminaisons des mots, comme dans le **Test IVbis** :

CRF (but : Gi | feature (*mot, $D_3(mot), D_2(mot), D_1(mot)$*)) $\rightarrow$ ResbisGi

**Test VII** : Ce test consiste à réaliser un nouvel apprentissage avec les spécifications suivantes :
CRF (but : L2 | feature (*mot, lemme, ResG0, ResG1, ResG2, ResG3*)) $\rightarrow$ ResL2

**Test VIIbis** : comme **VII** mais sans les lemmes :
CRF (but : L2 | feature (*mot, ResbisG0, ResbisG1, ResbisG2, ResbisG3*)) $\rightarrow$ ResbisL2

**Test VIII** : Dans ce test, le CRF final du **Test VII** est remplacé par des règles de composition symboliques sur les résultats ResGi.

**Test VIIIbis** : comme dans **VIII** mais où les règles de composition symboliques opèrent sur les résultats ResbisGi.

La Figure 3 montre le résultat de l'apprentissage de chaque composante indépendamment, et les deux méthodes possibles utilisées pour recomposer des étiquettes complètes :

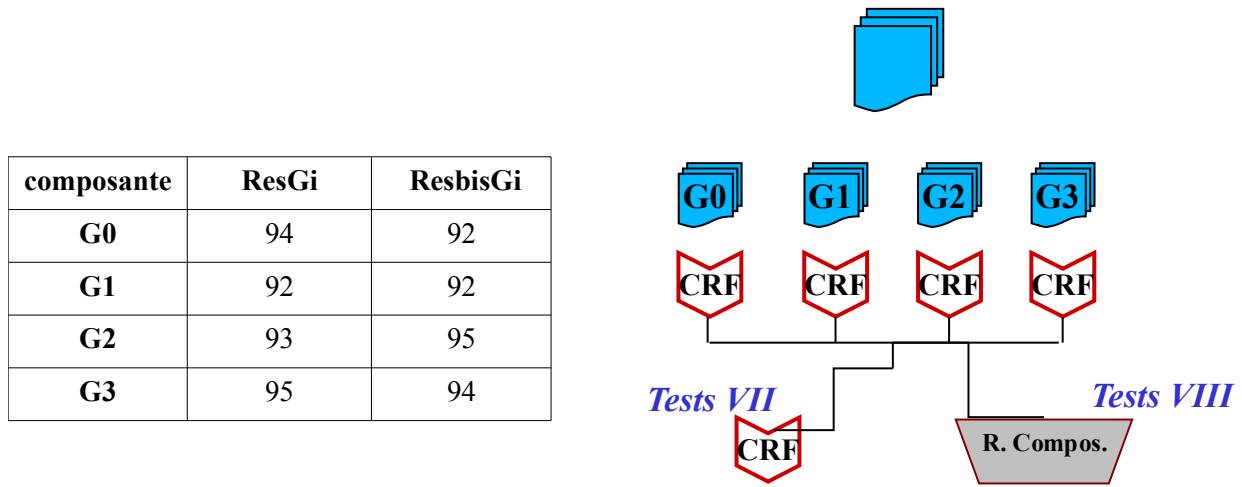

| composante | ResGi | ResbisGi |
|---|---|---|
| G0 | 94 | 92 |
| G1 | 92 | 92 |
| G2 | 93 | 95 |
| G3 | 95 | 94 |

*Figure 3 : Résultats de l'apprentissage des composantes et stratégies de combinaisons*

Les résultats de ces tests sont les suivants :

- **Test VII** : 89% de correction, **Test VIIbis** : 87,5% de correction ;
- **Test VIII** : 90% de correction, **Test VIIIbis** : 89,5% de correction.

L'apprentissage direct du niveau L2 (**Tests IV** et **IVbis**) n'est pas amélioré par la méthode de recomposition à base de CRF (**Tests VII** et **VIIbis**) mais elle l'est par la recomposition à base de règles (**Tests VIII** et **VIIIbis**). Notons aussi que, sans les lemmes mais avec les terminaisons des mots, certaines composantes sont bien apprises (notamment **G2**, la composante des nombres). Le **Test VIIIbis** illustre que, globalement, l'absence de lemmes peut être compensée par les terminaisons des mots, associées à une recomposition symbolique des étiquettes.

Par ailleurs, il faut remarquer que le temps d'apprentissage est considérablement réduit par cette stratégie de décomposition : le **Test VIII** ne prend que 75mn environ, contre 15h pour le **Test IV** (sur un PC standard), avec des résultats équivalents. Pour mieux mesurer le résultat de ces expériences, dans lesquelles les étiquettes obtenues sont souvent « partiellement correctes », il faudrait sans doute adapter la mesure de correction, qui ne prend pas en compte de ces subtilités.

## 4. Conclusion

Dans cet article, nous montrons qu'il est possible d'apprendre efficacement un étiqueteur morpho-syntaxique spécialisé sur un type de corpus particulier. Nous avons tout d'abord vu que les particularités de l'oral sont difficiles à énumérer sous forme de règles bien définies. Plutôt que de chercher à les caractériser, nous nous sommes donc contentés de fixer des conventions d'étiquetage pour les prendre en compte, et nous avons fait confiance à un système d'apprentissage automatique. L'approche que nous avons suivie prend les données telles qu'elles sont, sans éliminer aucune difficulté.

Notons qu'il n'est pas possible de comparer rigoureusement l'étiqueteur appris avec Cordial (ou avec tout autre étiqueteur) sur les mêmes données, parce que nous avons modifié la liste des étiquettes visées. Mais les performances des meilleurs étiqueteurs appris semblent tout à fait comparables à celles que Cordial obtient habituellement sur les corpus oraux.

L'intérêt des CRF pour réaliser cette tâche est qu'ils requièrent assez peu de paramétrages (il n'est pas nécessaire de comprendre toute la théorie des CRF pour utiliser CRF++) et assez peu de données de référence, mais qu'ils sont malgré tout suffisamment souples pour intégrer des connaissances linguistiques externes. Nous avons surtout utilisé ici la compréhension que nous avions des étiquettes pour nous focaliser sur l'apprentissage de sous-étiquettes plus simples et cohérentes. Si nous avions disposé d'un dictionnaire des étiquettes possibles pour chaque mot (ou chaque lemme), il aurait aussi été possible d'introduire des features qui prennent en compte cette nouvelle information (que nous n'avions pas ici).

Au vu de nos expériences, il semble difficile d'améliorer la correction des étiquettes apprises en jouant sur leur décomposition en sous-étiquettes plus simples. Mais cette décomposition permet d'améliorer considérablement le temps d'apprentissage. Pour confirmer ces résultats, cette stratégie devra bien sûr à l'avenir être testée dans d'autres contextes

## Références